\documentclass[sigconf]{acmart}

\AtBeginDocument{%
  \providecommand\BibTeX{{%
    \normalfont B\kern-0.5em{\scshape i\kern-0.25em b}\kern-0.8em\TeX}}}

\usepackage{graphicx}
\usepackage{subcaption}
\usepackage{siunitx}
\usepackage{natbib}

\global\long\def\ep{\mathbb{E}}

\copyrightyear{2020}
\acmYear{2020}
\setcopyright{acmlicensed}
\acmConference[ACM CHIL '20]{ACM Conference on Health, Inference, and Learning}{April 2--4, 2020}{Toronto, ON, Canada}
\acmBooktitle{ACM Conference on Health, Inference, and Learning (ACM CHIL '20), April 2--4, 2020, Toronto, ON, Canada}
\acmPrice{15.00}
\acmDOI{10.1145/3368555.3384451}
\acmISBN{978-1-4503-7046-2/20/04}

\begin{document}

\title{Population-aware Hierarchical Bayesian Domain Adaptation via Multi-component Invariant Learning}
\author{Vishwali Mhasawade}
\email{vishwalim@nyu.edu}
\affiliation{New York University}
\author{Nabeel Abdur Rehman}

\email{nabeel@nyu.edu}
\affiliation{New York University
}
\author{Rumi Chunara}
\email{rumi.chunara@nyu.edu}
\affiliation{New York University}


\begin{abstract}
While machine learning is rapidly being developed and deployed in settings such as influenza prediction, there are critical challenges in using data from one environment to predict in another due to variability in features. Even within disease labels there can be differences (e.g. ``fever'' may mean something different reported in a doctor's office versus in an online app). Moreover, models are often built on passive, observational data which contain different distributions of population subgroups (e.g. men or women). Thus, there are two forms of instability between environments in this observational transport problem. We first conceptualize the underlying causal structure of this problem in a health outcome prediction task. Based on sources of stability in the model, we posit that we can combine environment and population information in a novel population-aware hierarchical Bayesian domain adaptation framework that harnesses multiple invariant components through population attributes when needed. We study the conditions under which invariant learning fails, leading to reliance on the environment-specific attributes. Experimental results for an influenza prediction task on four datasets gathered from different contexts show the model can improve prediction in the case of largely unlabelled target data from a new environment and different constituent population, by harnessing both environment and population invariant information. The proposed approach will have significant impact in many social settings wherein \textit{who} the data comes from and \textit{how} it was generated, matters.
\end{abstract}





\maketitle
\section{Introduction}
\noindent 
Machine learning algorithms have the potential to significantly improve prediction efforts across critically important healthcare tasks. Yet, there are several issues that must be addressed before the potential of machine learning in health is broadly realized. While individual models are built on and may perform well on a select dataset from a specific environment (also called ``domain'' in the literature) and population (e.g. the population could be skewed towards younger people or other demographics depending on where it's sampled from), improving prediction in new datasets gathered in different contexts and simultaneously, from different constituent populations, is a clear challenge articulated by many health practitioners \cite{singh2019fair}.

First, standardization in health-related features is a significant problem. Variance in testing and billing practices \cite{mullainathan2019tested,pivovarov2014temporal} as well as differences in clinical case definitions \cite{ray2017predicting} from one environment to another present barriers for model transport. Accordingly, the same symptoms (features) can mean different things in different environments; ``fever'' may mean something different reported to a doctor at hospital $A$ versus hospital $B$, or to a doctor compared to through a smartphone app \cite{ray2017predicting,rehman2018domain}. This issue is becoming more pertinent as the number and types of data collection environments (from clinical data, to healthworker-facilitated data wherein healthworkers visit individuals' houses, record symptoms and take specimens, to citizen-science studies in which participants report symptoms and submit specimens directly \cite{goff2015surveillance,fragaszy2016cohort}) is rapidly increasing. In all cases, obtaining labels can be impractical; e.g. for influenza they would require costly and time-consuming laboratory tests. Another critical challenge is that models are often built on data from a particular population in an environment, and transporting results to a different population can be challenging if subgroups are differently represented in source and target populations (representation bias \cite{suresh2019framework}). These differences in data collection and demographic distributions make the problem of predicting infection in a dataset by using data gathered from different environments and populations challenging. We therefore address this unique problem of domain adaptation in the presence of representation bias. We study the problem via a simple, but important influenza prediction task.

The idea of transporting observational findings from source environment(s) to a target environment is essential in science and the concept has been well-defined on the basis that target environments can often differ from source environments. Furthermore, it can be expensive to generate labels in a new environment \cite{pearl2011transportability}. Methods have been proposed to exploit the causal structure of the data generating process in order to address certain domain adaptation problems, each relying on different assumptions. While some work has focused on identifying the invariant components to ensure robust transfer \cite{subbaswamy2018learning,magliacane2018domain}, work by Pearl and Bareinboim \cite{pearl2011transportability}  showed that identifying the mechanisms by which two environments differ can also be used to inform empirical learning and incorporation of local variations in a system. With this background, in this paper we address the problem of observational transport with \textit{both} environment differences and population representation bias. We do this by proposing a new hierarchical domain adaptation model that includes population attributes in the hierarchy in order to capture invariant information through these multiple components. The model then allows transfer of invariant information as well as learning information specific to a local environment \textit{when necessary}. We are able to propose a solution to this problem by harnessing research in health regarding population structure (invariance in population attributes) along with algorithmic innovation to design this novel approach.

To accomplish this goal in a principled way, we first represent the data generating process (DGP) for our task via a selection diagram. Besides explicitly illustrating variables (nodes) and the mechanisms by which the nodes are assigned a value (edges) that are relevant to the DGP and do not vary across environments, a selection diagram includes $S$-variables which localize the mechanisms where sources of unreliability in the DGP exist. We formalize this description and discuss the selection diagram for the task in this study in following sections. We highlight that modeling the DGP requires an understanding of health concepts \cite{pearl2011transportability}. Thus for the task considered here (influenza prediction from symptoms) in order to identify the invariant and variant components of the causal graph, we leverage health research which shows that 1) reports of symptoms in relation to infection status vary by the data collection mode, and 2) while the population represented in an observational sample can suffer from selection bias, disease risk can be stratified by population groups
\cite{chunara2015estimating,saria2010learning}. In societally-prescient problems such as health, attributes of whom the data is from (population demographics like age, gender) are commonly available, and it is understood that there are shared characteristics within these groups \cite{saria2010learning}. 

In sum, we specifically address a situation in which both environment and constituent population change from the source to target datasets; often the case in health prediction tasks. We use a simple but important task of influenza prediction from symptoms, and four real-world datasets representing a diverse set of environments and populations. Specific contributions are: 1) Formalizing the DGP between symptom reports and infection status, capturing sources of stability and of variance across environments (which we categorize into two: selection bias and feature instability); 2) A new domain/environment adaptation model for observational transport that accounts for instability in observed features as well as improves prediction on population subgroups even when not well represented in a particular dataset, through sharing invariant population characteristics in multiple components \textit{as needed} (when a population subgroup is not well-represented in the target environment or its characteristic is different from that in other data); 3) Demonstrating the model on real-world data, showing significant improvement in prediction of infection on largely unlabelled target dataset overall \textit{and} by population subgroups in comparison with several relevant baselines.

\section{Notation and Problem Setting}
We consider source datasets from multiple environments \\
$\mathcal 
{D}_{e} := \{(x_{i}^{e},y_{i}^{e},a_{i}^{e},g_{i}^{e})\}_{i=1}^{n_{e}}$ where $e \in E$ ($E$ comprises of all the source environments) and a single target dataset
\begin{align*}
    \mathcal{D}_{t} :=  \{(x_{i}^{t},y_{i}^{t},a_{i}^{t},g_{i}^{t})\}_{i=1}^{k} \bigcup \{(x_{i}^{t},a_{i}^{t},g_{i}^{t})\}_{i=k+1}^{n_{t}}
\end{align*}
where $k << n_{t}$; $t \in T$. For the target dataset we have limited number of labeled samples $\left(k\right)$ whereas for the source datasets all the samples are labeled. $L$ denotes all the datasets: source as well as the target $\left(L = E \cup T\right)$. Sets of variables are denoted by italicized capital letters whereas lowercase letters are used for their individual assignments.  

$Y$ denotes the presence $\left(y =1\right)$ or absence $\left(y=0\right)$) of the influenza virus. Age of the individual is represented using $A$, and categorized by common epidemiological groups: age 0-4, age 5-15, age 16-44, age 45-64, age 65+. Similarly, $G$ represents gender (male or female). The demographic attributes ($A$ and $G$, but can be expanded to other demographic attributes where possible) are together represented as $D$; $D = \{A,G\}$. $X$ is the feature vector representing presence of the symptoms: fever, cough, muscle pain and sorethroat. Here $x$ is a 4-dimensional binary vector representing the symptoms that an individual has (if an individual $i$ has fever and sorethroat but no cough and muscle pain; the feature vector looks like $x_i = \{1,0,0,1\}$). We consider subgroups in the data to be the specific demographic populations of interest belonging to a specific gender and age group $\mathcal{D}_{a,g} = \{(X,Y) \mid A=a, G=g \}$. The task is to predict the value of $Y$ for each of the subgroups $\mathcal{D}_{a,g}$ from the symptom information $X$. This can be formalized as:
\begin{equation*}
   \min_{\forall a, g} R^t\left(f\left(X^t,\theta^t\right)\right) + \sum_{e} R^e\left(f\left(X^e,\theta^e\right)\right)
\end{equation*}

\noindent We aim to learn classifier $f\left(X^t,\theta^t\right)$ for the target dataset $\mathcal{D}_t$ parameterized by $\theta^t$ for each of the demographic subgroups ($\mathcal{D}_{a,g}$) that minimizes empirical risk $R^t$ while minimizing total risk across the source environments $R^e$ as well. It should be noted that the probability distribution of the target environment across population subgroups $P_t\left(X,Y\mid D\right)$ may not be uniform. Hence, the resulting $f\left(X^t,\theta^t\right)$ cannot be assumed to be the same across all subgroups.

\section{Related Work}
\textbf{Influenza Prediction}
Influenza is a global threat, affecting countries worldwide with considerable morbidity and mortality \cite{reich2019collaborative}. Globally, annual epidemics are estimated to result in about 3 to 5 million cases of severe illness, and about 290,000 to 650,000 respiratory deaths \cite{world2018influenza}. With the possibility of global pandemics looming, improving prediction of influenza is a continuing central priority of global health preparedness efforts. Predicting from symptoms in single datasets have used regression models \cite{monto2000clinical}, typically examining specific case definitions (sets of syndromic features). Machine learning approaches have enabled wider feature space examination \cite{pineda2015comparison}. While it is understood that health-related features can vary from hospital to hospital \cite{wiens2014study}, influenza data sources incur even more diversity as passive observations are collected via varied sources including syndromic surveillance systems, Internet apps, and health worker based studies. Also, generating labels is difficult and costly (requires laboratory testing). Recent work has shown that domain adaptation can be useful for prediction from symptom data sets obtained via these different environments \cite{rehman2018domain}. While epidemiological study has indicated that there are disparities in risk by age group and gender for disease in general, and influenza specifically \cite{bansal2010shifting}, prediction approaches that harness population attribute differences are an important gap in disease prediction models. \\

\noindent \textbf{Observational transport.} 
Observational transport refers to the transport of causal relationships across environments in which only passive observations can be collected \cite{pearl2011transportability}. The simple idea indicates that causal knowledge shows which mechanisms remain invariant under change. Accordingly, some work has used causal diagrams or feature selection methods to determine invariant relations in the source environment that can be transferred to the target environment, isolating the set of features which can be conditioned on to eliminate instabilities in the data generating process \cite{subbaswamy2018learning,mooij2016joint,magliacane2018domain}. Though it should be noted that early work goes on to state that the causal relation to be transported can be learned from invariant components and \textit{variant} components from both the source and target environments, depending on the DGP \cite{pearl2011transportability}. Here, we use this idea to allow trade-off between invariant characteristics across environments and empirical re-learning of relationships from each local environment, depending on which populations are represented in a dataset. In other words, we transmit invariant information through multiple population components, and use variant information \textit{as necessary}, addressing the problem of different population subgroup representation in observational data.\\

\noindent \textbf{Multi-source domain adaptation and hierarchical modeling}. 
Domain adaptation is focused on improving performance for a target data set, in situations where the environment of the target data is different from the that of the source(s) from which information is transferred. Another approach to learning from multiple sources by pooling and analyzing multi-site datasets includes transforming the source and target feature spaces to correct any distributional shift in the data \cite{zhou2018statistical}. Prior work have also leveraged multiple source datasets to increase the amount of information learned \cite{guo2018multi}. This task has also been formulated from a causal view \cite{mooij2016joint}, where the posterior of the target is a weighted average of the source datasets. The ``Frustratingly Easy Domain Adaptation'' method is notable for simplicity and good performance on text data \cite{daume2009frustratingly} and is equivalent to hierarchical domain adaptation \cite{finkel2009hierarchical} (except it explicitly ties parameters across environments). Hierarchical approaches, which have primarily been developed in natural language processing, in contrast allow hyperparameters to be separated across environments; each environment has its own environment-specific parameter for each feature which the model links via a hierarchical Bayesian global prior instead of a constant prior. This prior encourages features to have similar weights across environments unless there is good contrary evidence. This supports the goal of this work, to combine environment information as needed (unless the population represented in the local environment is much different than in other environments). Hierarchical Bayesian frameworks are a more principled approach for transfer learning, compared to approaches which learn parameters of each task/distribution independently and smooth parameters of tasks with more information towards coarser-grained ones \cite{carlin2010bayes}. In this work we advance this idea by creating a novel multi-level, multi-component hierarchy, as well as by the idea of incorporating  population-attribute invariance as part of the hierarchy.

\section{Proposed Approach}
\subsection{Assumptions}
Here we describe the assumptions that ensure our problem is well-posed. The main assumption is that the data generating process is known and can be represented via a graphical causal diagram (helps to identify the information that can be transported \cite{pearl2011transportability}). We adapt the definition of a selection diagram which is previously defined \cite{pearl2002causality,pearl2011transportability} to clearly delineate different types of change mechanisms. \\

\noindent \textbf{Definition 1 (Selection diagram).} \textit{A selection diagram is a probabilistic causal model (as defined in \cite{pearl2002causality}) augmented with auxiliary selection variables $S$ (denoted by square nodes, which denote places of instability in the DGP) comprising of two types; $S = \{S^*,\Tilde{S}\}$. An $S^*$ variable can point to any observed variable.  $S^* \rightarrow X$ denotes that the mechanism of assigning value to $X$ changes across environments. The other type of selection variable $\Tilde{S}$ represents a selection bias. Thus an edge from $X$ to $\Tilde{S}$ $\left(X \rightarrow \Tilde{S}\right)$ denotes a non-random selection of individuals, groups or data for variable $X$.} \label{definition:selection}\\

\begin{figure}[b]
    \centering
    \includegraphics[scale = 0.40]{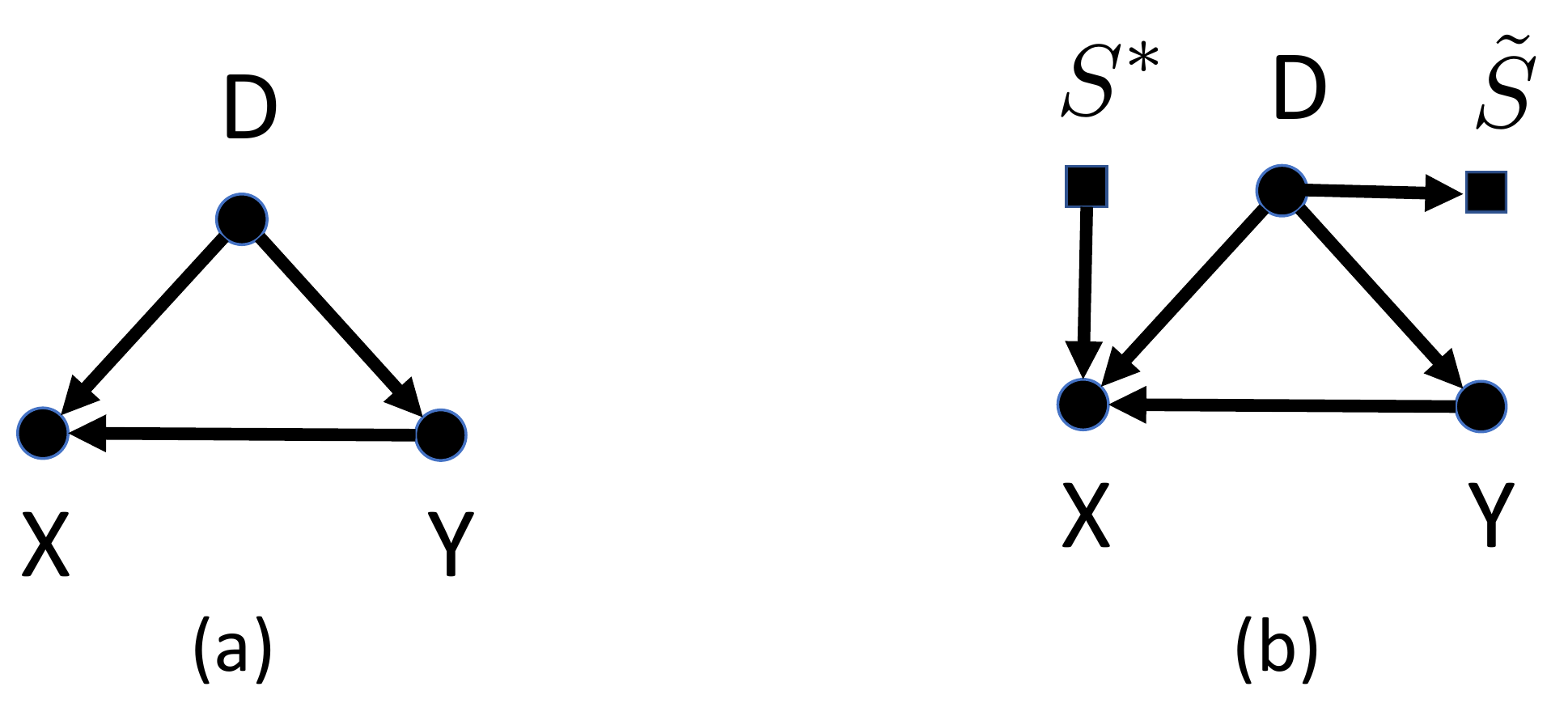}
    \caption{(a) Causal diagram, (b) Selection diagram representing the differences in the data-generating process.}
    \label{fig:causal_diagram}
\end{figure}

\noindent We can now formalize the causal and selection diagrams (Figure \ref{fig:causal_diagram}) for our setting (prediction of influenza infection from symptoms) based on prior knowledge and research in health. Along with the system variables: virus $\left(Y\right)$, symptoms $\left(X\right)$ and demographic attributes $\left(D\right)$ of age and gender, we also have the selection variables $\left( S = \{S^{*},\Tilde{S}\}\right)$ which denote differences in the data-generating process across environments through instability in observed variables and selection bias. The symptoms that result are generally shaped by infection status \cite{CDC_Flu_Symp}, thus we have $Y \rightarrow X$. Population demographic attributes also can affect symptoms reported, $X$, and susceptibility to infection by the virus, $Y$ (for example, symptoms common in young versus older people can vary; $X \leftarrow D, D \rightarrow Y \rightarrow X$) \cite{chunara2015estimating}. Now, we consider the parts of the data-generating process that vary across environments. The data collection environment (here, for example citizen science or health-worker facilitated) affects $P\left(X \mid Y\right)$ (specifically, it is known that symptoms reported via citizen science are less specific than in a hospital, for example) \cite{ray2017predicting}. Thus the collection source introduces differences in the manner in which $P\left(X\mid Y\right)$ is observed across environments and there is a selection variable pointing towards $X$; $\left(S^* \rightarrow X \right)$. The absence of a selection variable pointing at $D$ and $Y$ indicates that the mechanism of assigning values to these variables is the same across environments (which makes sense intuitively, as demographic variables, e.g. man or woman, do not change or have different meanings in the different environments, nor does the process for obtaining flu infection status which is performed by laboratory confirmation in all cases). Finally, there is a selection bias associated with population demographic attributes. The proportion of individuals in each of the subgroups commonly varies across environments based on observational sampling (it is rare to have a representative distribution in a population sample unless an experiment is designed in advance and specific groups are recruited); $P_e \left(X,Y\mid D\right) \neq P_t \left(X,Y \mid D\right)$. Thus there is an edge from $D$ to $\tilde{S}$. We now state the assumptions that help to formulate observational transport for this causal structure.\\

\noindent\textbf{Assumption 1.} \textit{Let $\mathcal{G}$ be a causal graph with variables V consisting of the system variables $\mathcal{I} = \{ X,Y,D\}$ and the selection variables $\mathcal J = \{S^\ast\}$ }. 
\begin{enumerate}
    \item \textit{No system variable directly causes any selection variable} \\
    $\left( \forall j \in \mathcal{J}, \forall i \in \mathcal{I}: i \rightarrow j \notin G \right)$.
    \item \textit{No system variable is confounded by any selection variable ($S^\ast , \tilde{S}$).}\\
\end{enumerate}

\noindent\textbf{Assumption 2.} \textit{Let $\mathcal{G}$ be a causal graph with variables V consisting of the system variables $\mathcal{I} = \{ X,Y,D\}$ and the selection variables $\mathcal S = \{S^\ast,\tilde{S}\}$ and $P(V)$ be the corresponding distribution on $V$. }
\begin{enumerate}
    \item \textit{The distribution $P\left(V\right)$ is Markov and faithful with respect to $\mathcal{G}$.}
    \item \textit{$S$} has no direct effect on $Y$ $\left(S \rightarrow Y \notin G\right)$)

\end{enumerate}

\subsection{Observational transport across environments}
\noindent Motivated by the approach stated in \cite{pearl2011transportability} we aim to leverage a statistical relation, $R\left(P\right)$ to be learned from source environment(s) (characterized by probability distribution $P$) and transfer it to another (target) environment, $R\left(P^{\ast}\right)$, (characterized by probability distribution $P^\ast$)  particularly when gaining complete information about that relationship in the target environment is costly. The definition of observational transportability in \cite{pearl2011transportability} (Definition 5), asserts that the relation to be transported has to be constructed from the source data as well as observations from the target data. As there is no control on the data-generating process (no intervention on any of the system variables, in contrast to experimental data) we cannot use \textit{do}-calculus for formalizing the causal relation, and instead must use conditional independencies to understand the relationship between the outcome, $Y$ and features $X$, by obtaining the joint probability distribution $P^{\ast}\left(X,Y,D\right)$. In the following section we identify invariant parts of this relation (which can be learned in combination with the source environment), and transferred as well as the target environment-specific relations (variant components) to be learned directly from the target dataset.

\subsection{Multi-component invariant transfer}
Having knowledge of the data-generating process via the graphical casual model $\mathcal{G}$, we identify the invariant conditional distributions that can be transferred from the source environment $\left(\mathcal{D}_e\right)$ to the target environment $\left(\mathcal{D}_t\right)$. 

Indeed, according to the causal diagram in Figure \ref{fig:causal_diagram}b, we do not find a set of features $\left(X\right)$ that d-separates $S$ and $Y$, $S  \not\!\perp\!\!\!\perp Y \mid X$. However, we do notice that $S \perp \!\!\! \perp Y \mid D$; the invariant information $P\left(Y|D\right)$ can be transferred across the environments. This follows from the fact the different demographic subgroups of the population share characteristics; for example, babies are known to be susceptible to certain infections as opposed to older people; strengthening the fact that the conditional distribution $P\left(Y|D\right)$ can be transferred across environments. However, we do need to learn $P^\ast\left(Y \mid X,D\right)$ for the target dataset since $S  \not\!\perp\!\!\!\perp Y \mid X,D$. We therefore present an approach to learn the environment specific component $\left(P^\ast\left(Y\mid X\right)\right)$\footnote{$P^\ast\left(Y \mid X \right) = \sum_D P^\ast\left(X \mid Y,D\right) \cdot \frac{P\left(Y \mid D\right)}{P^\ast\left(X\mid D\right)}$} as well as the population invariance $P\left(Y\mid D\right)$ from shared characteristics.\\

\subsection{Formal framework of the undirected hierarchical multi-source Bayesian approach}
\begin{figure}
    \centering
    \includegraphics[width=0.9\linewidth]{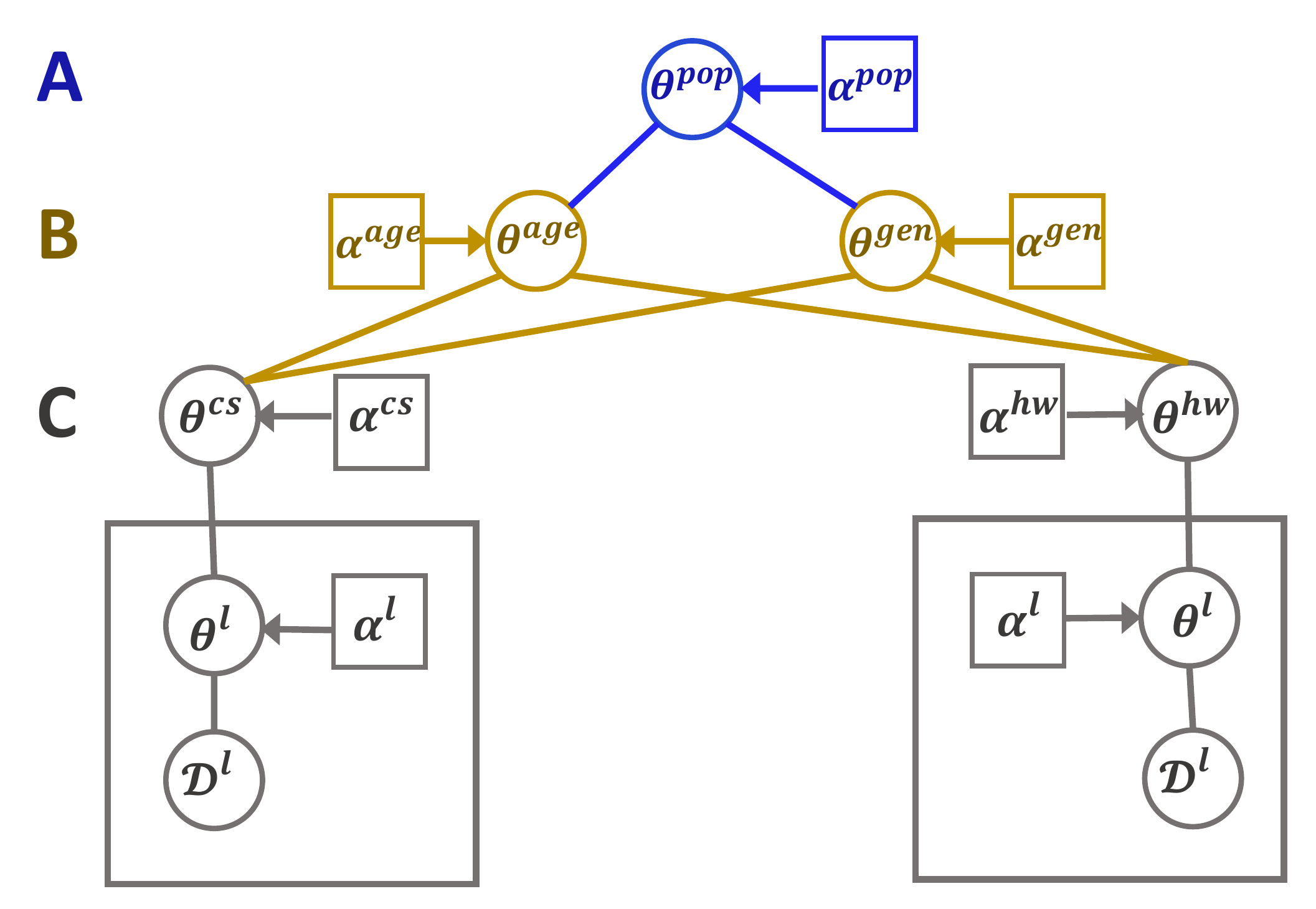}
    \caption{Population-aware hierarchical model; $\theta$ parameters at different nodes, $\mathcal{D}$ different data sets, $\alpha$ the priors. (A): Root level that represents invariant information across all data, (B): population parameters and information invariant to population-attributes $\left(\mathit{age}\right)$ and $\left(gender\right)$, (C): data set and environment-specific parameters and information ($\mathit{cs}$  for citizen science and $\mathit{hw}$ for healthworker facilitated datasets).}
    \label{fig:popware_model}
\end{figure}

Having identified the sources of variability and stability, we now can describe details of the model specific domain adaptation approach which enables learning  $P^\ast\left(Y \mid X\right)$ and $P\left(Y\mid D\right)$, as described in the previous section. In the framework, the lowest level of the hierarchy represents the datasets (within each environment, in our case, citizen science or health-worker facilitated), $l\in L $, for each of which we have the labeled data $\mathcal{D}_l$ of the dataset $l$ as shown in Figure \ref{fig:popware_model}. As in all Bayesian settings, the dataset parameters $\theta^{l}$ should represent the data $\mathcal{D}_l$ well. Here, $\theta^{l}$ are influenced by the environment-specific parameters $\left(\theta^{c}\right)$; $\theta^l$ are generated according to $P\left(\theta^{l} \mid \theta^{c}\right)$, where  $c \in {C}$ is the collection mode and $\theta^c = \{ \theta^{\mathit{cs}},\theta^{\mathit{hw}} \}$ where $\theta^{\mathit{cs}}$ represents the parameters for the citizen-science collection mode and $\theta^{\mathit{hw}}$ represents the parameters for the health-worker supported collection mode. In the undirected hierarchical model we allow the environment specific parameters to have multiple parents and learn all parameters simultaneously. Accordingly, the environment  parameters are generated according to the distribution $P\left(\theta^{c} \mid \theta^{a}, \theta^{g} \right)$. Here, we explicitly represent the population parameters; $\theta^{a}$ for ${a} \in {A}$, the different age group categories, and $\theta^{g}$ for genders ${g} \in G$, $\theta^{d} = \{\theta^{a},\theta^{g} \}$ and $ d \in D $. The model thus learns the invariant component parameters ($\theta^d$) for the different demographic subgroups (ages 0-4, 5-15, 16-44, 45-64, 65+, males, females). Population \\
parameters $\theta^{a}$ and $\theta^{g}$ have the root parameter $\theta^{\mathit{pop}}$ as the parent, which represents invariant information across all of the datasets, environments and population attributes, $P\left(\theta^{\mathit{pop}}  \mid \theta^{\mathit{par\left(pop\right)}}\right) \equiv P\left(\theta^{\mathit{pop}}\right)$. Then, the joint distribution is:\\
\begin{align*}
    P\left(X,Y, \theta \right) &=   \prod_{l \in {L}} P\left(\mathcal{D}_l\mid \theta^{l}\right) \times 
    \prod_{l \in {L}} P\left(\theta^{l}\mid \theta^{c}\right)  \times
    \prod_{c \in {C}} P\left(\theta^{c} \mid  \theta^{a},\theta^{g}\right) 
    \\
    &\times \prod_{a \in {A}}
    P\left(\theta^{a} \mid \theta^{\mathit{pop}} \right) \times
    \prod_{g \in G}
    P\left(\theta^{g}\mid \theta^{\mathit{pop}}\right) \times P\left(\theta^{\mathit{pop}}\right)
\end{align*}

We also study the conditions under which the invariant component parameters $\left(\theta^d\right)$ do not completely represent the information for a subgroup in which case the environment specific parameters $\left(\theta^l\right)$ help; thus explicating the conditions under which the invariant information is useful, and when environment-specific information should be utilized.

\subsection{Hierarchy priors}
For all parameters we use independent priors, computed based on symptom predictivity for each age group and gender. The inclusion of data dependent priors in Bayesian learning has been explored to incorporate domain knowledge into the posterior distribution of parameters \cite{darnieder2011bayesian}. 
For population-aware modeling, data-informed prior distributions are important because the distributions from each dataset are particular to the study, and thus capturing this information adds more information to the analysis than improper or vague priors (e.g. for a sample wherein one demographic group is under-represented), also motivates the multiple parents in the hierarchy.
In contrast, using just the root prior for estimating the posterior ignores the demographic information available.  
Therefore, we use an empirical Bayes approach to specify weakly informative priors, centered around the estimates of the model parameters \cite{van2017prior}. Root parameters are centered on the cumulative data since the root parameter captures environment invariant information.

\subsection{Model steps}
 First, we use a probabilistic framework to jointly learn each parameter based on all levels of the hierarchy. We use a maximum a-posteriori parameter estimate instead of the full posterior for the joint distribution, which would be computationally intractable. We use a formulation, proposed in \cite{elidan2012convex} that is amenable to standard optimization techniques, resulting in the objective:

\begin{equation}
\begin{aligned}
   F_{\mathit{objective}} = -\sum_{l \in L} \bigg[ \sum_{j} (f_j  + \lambda) \cdot \theta_{j}^{l}
   - \log \sum_{k} \exp(\theta_{k}^{l})\bigg] \\
   + \beta\sum_{n \in \mathit{Nodes}} \mbox{Div}(\theta^{n}, \theta^{\mathit{par}(n)})
    \label{eqn:obj}
\end{aligned}
\end{equation}

\noindent For dataset $l$, $\theta_{j}^{l}$ denotes the parameter for  symptom $j$. From a specific dataset's parameter space, $k$ represents individual symptoms. $f_{j}$ is a statistical measure of the symptom $j$ in the dataset, in this case the proportion of the particular symptom resulting in a positive influenza virus (i.e. the positive predictive value). $\mathit{Nodes}$ is the set of all nodes in the hierarchy (here, ${L} \cup {C} \cup {A} \cup G$).  Regularizing parameter $\lambda$ was chosen as 1 to allow Laplacian smoothing. The function $\mbox{Div}(\theta^{n},\theta^{\mathit{par(n)}})$ is a divergence (L2 norm used) over the child and parent parameters that encourages child parameters ($\theta^n$) to be influenced by parent parameters ($\theta^{\mathit{par}(n)}$), and allows a child parameter to be closely linked to more than one parent. The weight $\beta$ represents the influence  between node parameters and node parent parameters. Based on hyperparameter tuning, a value of 0.2 for $\beta$ was used in all experiments. For  objective function optimization we use Powell's method \cite{fletcher1963rapidly}.

Second, we learn the influence $\left(\gamma\right)$  of each parent on a particular dataset (child node). This is necessary since we need to learn $P^*\left(Y\mid  X,D\right)$ for the target dataset as observed from the causal structure. We provide a mechanism to learn that as follows:
\begin{align*}
    y_i^{\left(l,a,g\right)} = \gamma_{0} + \gamma_{1}\left(\theta^l \cdot x_i^{\left(l,a,g\right)}\right) + \gamma_{2}\left(\theta^{a} \cdot x_i^{\left(l,a,g\right)}\right) + \gamma_{3}\left(\theta^{g} \cdot x_i^{\left(l,a,g\right)}\right)
\end{align*}
The weights $\gamma_0, \gamma_1, \gamma_2, \gamma_3$ are estimated from a non-linear least square regression; the information from the different parents and the dataset can only be positive and hence we restrict the weights to be positive. This enables the model to give more weight to one level of the hierarchy when needed. In other words, how much demographic-invariant or environment-specific information is needed depends upon how much information is in a given dataset. For each of the subgroups a different classifier is learned based on the preferences of the subgroup. The reason for learning the weights for the different levels for each dataset independently is that each dataset would require different amounts of information from the demographic-specific and the environment-specific parameters, depending upon the demographic distribution of the sample in that dataset as well as the environment.

\subsection{Licensing conditions for the use of invariant representations} \label{conditions}

To understand the cases under which the invariant representations captured by $\theta^{a},\theta^{g}$ fail to capture information for a specific subgroup, and local data must be used, we analyze information at the demographic subgroup level. The model structure consists of different hierarchies wherein each hierarchical level learns invariant information. This implies that invariant information learned by the higher levels is invariant across environments as compared to the leaf nodes in which data-specific information is learned. We begin by describing the conditions on which information is evaluated.\\

\noindent \textbf{Definition 3.} \textit{Let}\\
$P_{\mathit{diff}}\left(X \mid Y=y\right) =\big | P\left(X=1 \mid Y=y \right) - P\left (X=0 \mid Y=y\right) \big |$ \\
\textit{ be the difference of conditional probabilities of X (symptoms) given Y equal to y.}\\

\noindent \textbf{Definition 4.} \textit{Let} \\
$\delta_{\mathcal{D}} = \ep_{x \in \mathcal{D}, y \in \mathcal{D}} \left[  P_{\mathit{diff}}\left(X \mid Y=1,A=a,G=g \right )\right]$ \\
\textit{be the expectation of $P_{\mathit{diff}}$ over the symptoms for the subgroup $\mathcal{D}_{a,g}$ of the dataset $\mathcal{D}$. Similarly we define $\delta_{\mathit{pop}}$ to be the expectation of $P_{\mathit{diff}}$ over the symptoms for the population subgroup \quad $\cup \mathcal{D}_{a,g}^l$ comprising of the subgroups from all the environments ($l \in L$).}\\

\noindent \textbf{Theorem 1.} \textit{The parameters $\theta$  for a subgroup ($\mathcal{D}_{a,g}$) of a dataset ($\mathcal{D}$) depends on the $\delta_{\mathcal{D}}$ and the conditional probability $P_{\mathit{pop}_{a,g}}\left(Y\right) = P\left(Y=1 \mid l=\mathit{pop},A=a,G=g \right)$ for the entire population comprising of the subgroups from the individual environments and the conditional probability $P_{\mathcal{D}_{a,g}}(Y) = P\left(Y=1 \mid l=\mathcal{D},A=a,G=g\right)$ for the subgroup of the specific dataset. }

\begin{equation*}
  \theta=\left\{
  \begin{array}{@{}ll@{}}
    \theta^l, & \text{if}\ \delta_{\mathcal{D}} < \delta_{\mathit{pop}} \\
    \theta^l, & \text{if}\ P_{\mathcal{D}_{a,g}}\left(Y\right) - P_{\mathit{pop}_{a,g}}\left(Y\right) \approx 1\\
    
    \theta^{d}, &\text{otherwise}
  \end{array}\right.
\end{equation*}
\begin{proof}[Proof Sketch] (full proof in Appendix)\\
a) We make use of the information function $I=-\left[\log \left(P_h\right)\right]$ which represents the information present about event $h$.  If $\delta_{\mathcal{D}} < \delta_{\mathit{pop}}$ then $P\left(X=1\mid Y=1,d=l\right)< P\left (X=1 \mid Y=1,d=pop\right)$ (this condition is explained in the proof in the appendix). Since $I$ is a monotonically decreasing function, $I_d > I_{\mathit{pop}}$. Since the specific dataset has more information, the dataset specific parameters are used instead of using the invariant parameters learned over all the global population.\\
\noindent b) $P_{\mathcal{D}_{a,g}}\left(Y\right) - P_{\mathit{pop}_{a,g}}(Y) \approx 1$ if $P_{\mathcal{D}_{a,g}}(Y) \approx 1$ and $P_{\mathit{pop}_{a,g}}(Y) \approx 0$. This means that the specific subgroup (${a,g}$) is over represented in the specific dataset $l$ but we do not have much information about the specific subgroup from the invariant global representation since it is underrepresented in the global population.
\end{proof}
\noindent The conditions determine the cases in which spurious relations could be picked up by the invariant component representations $\theta^d$ and hence the data-specific parameters $\theta^l$ better represent the relations persistent in the specific dataset. The theorem states the conditions under which the invariant component representations $\theta^d$ will be used and when we need to rely on the data-specific parameters $\theta^l$ to capture the relations for a specific subgroup of the dataset.

\section{Data}
\begin{table*}[t]
\centering
\caption{Summary of dataset details.}
\begin{tabular}{c c c c }
\toprule
	Study & Location & Observations (positive) & Collection Type \\
\midrule
Goviral & Northeast United States & 520 (291) & Citizen Science \\
    Fluwatch & England, United Kingdom & 915 (567) & Citizen Science  \\
    Hongkong & Hong Kong & 4954 (1471) & Healthworker Facilitated  \\
    Hutterite & Alberta, Canada & 1281 (787) & Healthworker Facilitated \\
\bottomrule
\end{tabular}
\label{tab:data}
\end{table*}

\begin{table}[t]
 \centering
 \caption {AUC for flu prediction task (with 20\% labeled data from target), \textbf{bold} values correspond to best performing model.}
 \begin{tabular}{l c c c c}
 \toprule

  & Goviral & Fluwatch & Hongkong & Huttterite \\
    \midrule
   {TR}  & {0.594} & {0.584} &  {0.865} &  {0.712}  \\
   
   {LR}  & {0.585}  & {0.490} & {0.914}  & {0.706} \\

    FEDA & {0.588} & {0.521} &  {0.806}  & {0.651}\\
    
    FEDA+pop & 0.500 & 0.442 & 0.727 & 0.582 \\
   
   Hier & 0.645 & 0.546 & 0.881 & 0.680 \\
   
   Hier+pop  & {\textbf{0.744}} & {\textbf{0.754}} & {\textbf{0.919}} & {\textbf{0.814 }}\\
\bottomrule
\end{tabular}
   \label{tab:result}
 \end{table}
 
\begin{figure}[b]
\centering
\includegraphics[scale = 0.36]{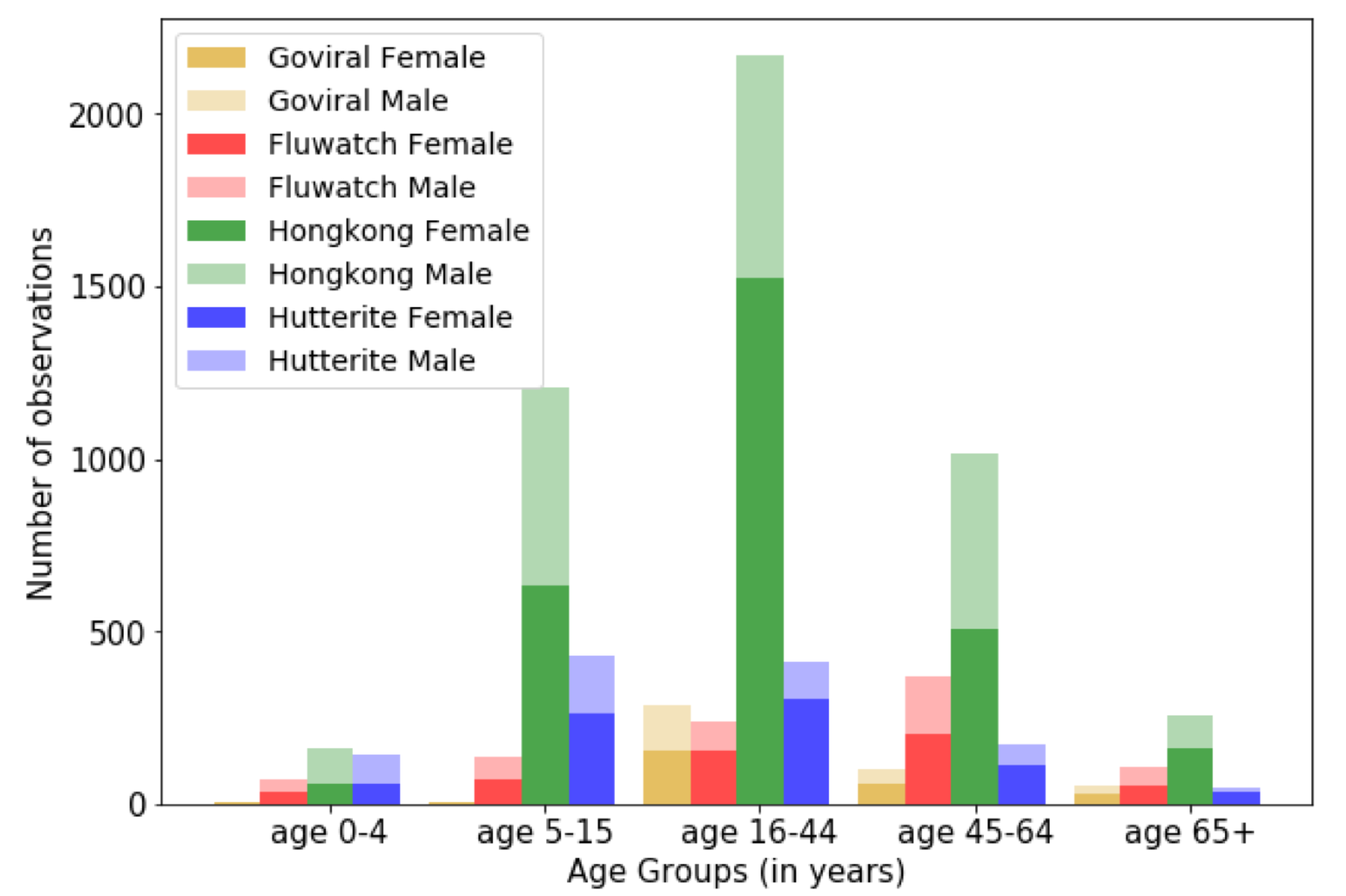}
\caption{Demographic distributions in datasets. Darker shade of color denotes the number of females in the particular age group, lighter shade denotes number of males.}
\label{fig:demo}
\end{figure}

Each dataset includes symptoms from individuals ($X$), laboratory confirmation of type of influenza virus they had (if any) ($Y$), and age and gender ($D)$ of each person as example population attributes. Attributes of the datasets are summarized in Table \ref{tab:data}, while the breakdown of positive and negative observations across demographics is shown in Figure \ref{fig:demo}. Through these differing study designs, one can see how the features may differ based on the stage of illness and included populations (e.g. those who have healthworkers come visit, versus those stay at home and may be sick but well enough to report on their own). As well, the difference in underlying populations illustrates the need for combining data in a principled way and accounting for these differences in the underlying sample composition. It should be emphasized that each of the datasets have a varied composition in terms of total number of observations and population demographics (Appendix Figure 1). We choose to use them all without any pre-processing, as these demonstrate real data set differences and will indicate model performance in such real-world situations. Indeed, population subgroups are not equally represented across all datasets. Goviral and Hongkong have the highest proportion of observations in the age group of 16-44, Fluwatch has the highest proportion of observations across the age group 45-64 while Hutterite has the highest proportion of observations in the age group of 5-15. The first two studies we classify as ``citizen science'' as they involve individuals self-reporting symptoms themselves from home, and taking their own nasal specimens for microbiological testing. The next two studies we classify as ``healthcare worker'' as they involve a trained worker visiting the individual, recording their symptoms according to a set criteria, and taking a nasal specimen from the participant. These studies also generally involved people more likely to be infectious \cite{cowling2010comparative}. Below we summarize the study design and context behind each dataset, including how the data was collected, while references are provided for the full papers describing all details.\\

\noindent The \textbf{GoViral} data comes from volunteers who self-reported symptoms online and also mailed in bio-specimens for laboratory confirmation of illness. These participants thus were never visited at home or in person at all. Volunteers were recruited, given a kit (collection materials and customized instructions), instructed to report their symptoms weekly, and when sick with cold or flu-like symptoms, requested to collect a nasal swab \cite{goff2015surveillance}. Periodic reminders were sent over email. Data from 2013--2017 is included.\\

\noindent \textbf{FluWatch} was a study in the United Kingdom, consisting of households which were recruited from registers of 146 volunteer general practices across England in seasonal and pandemic influenza over five successive cohorts from 2006--2011 \cite{fragaszy2016cohort}. Individuals participated from their home. In addition to a baseline visit by a nurse, households received participant packs containing paper illness diaries, thermometers and nasal swab kits including instructions on their use and the viral transport medium to be stored in the refrigerator. While participants would generate specimens and illness reports on their own, they were reminded every week via automated phone calls. \\

\noindent The \textbf{Hong Kong} study was focused on measuring infection in people who had household members who were already confirmed as sick. Household contacts of index patients (people who had come to the hospital and were confirmed to be sick) were followed in July and August 2009. These contacts live in close proximity with people who's illness was severe enough to take them to hospital, indicating a high risk for infection. Household members of 99 patients who tested positive for influenza A virus on rapid diagnostic testing were visited at their homes and swabs collected from household members by health workers over multiple weeks \cite{cowling2010comparative}.\\

\noindent Data is also included from a study wherein nurses sampled people in \textbf{Hutterite} colonies in Alberta, Canada. The Hutterites are an ethno-religious group that tend to live together in colonies that are relatively isolated from towns and cities, and therefore are interesting places to examine respiratory infection prevalence given their self-contained nature.  Data is from Dec. 2008 to June 2009 \cite{loeb2010effect}.\\

\section{Experiments}

\begin{figure*}
    \centering
    \begin{subfigure}{0.5\textwidth}
    \includegraphics[width=1.1\linewidth]{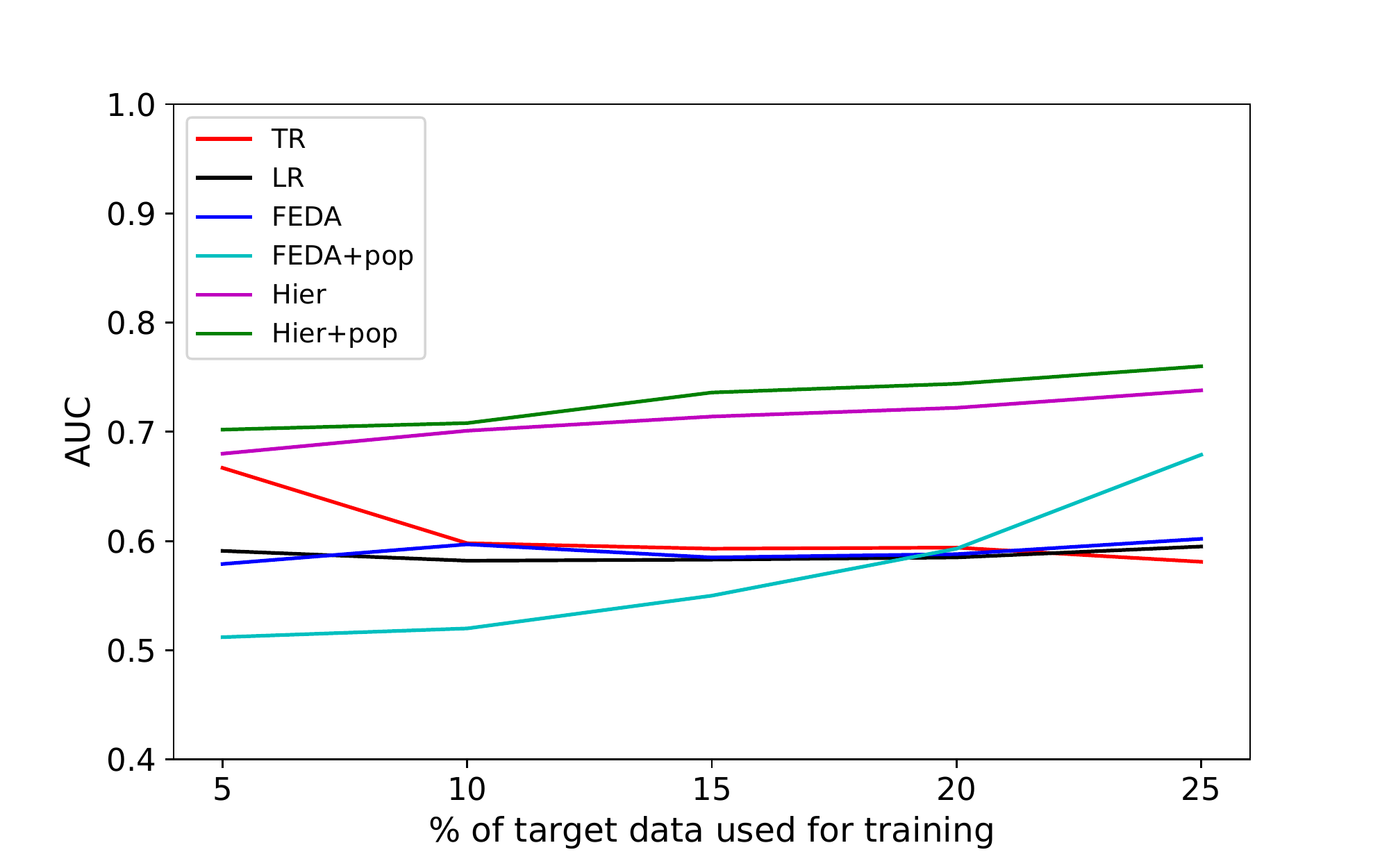}
    \caption{Goviral}
    \end{subfigure}%
    \begin{subfigure}{0.5\textwidth}
    \includegraphics[width=1.1\linewidth]{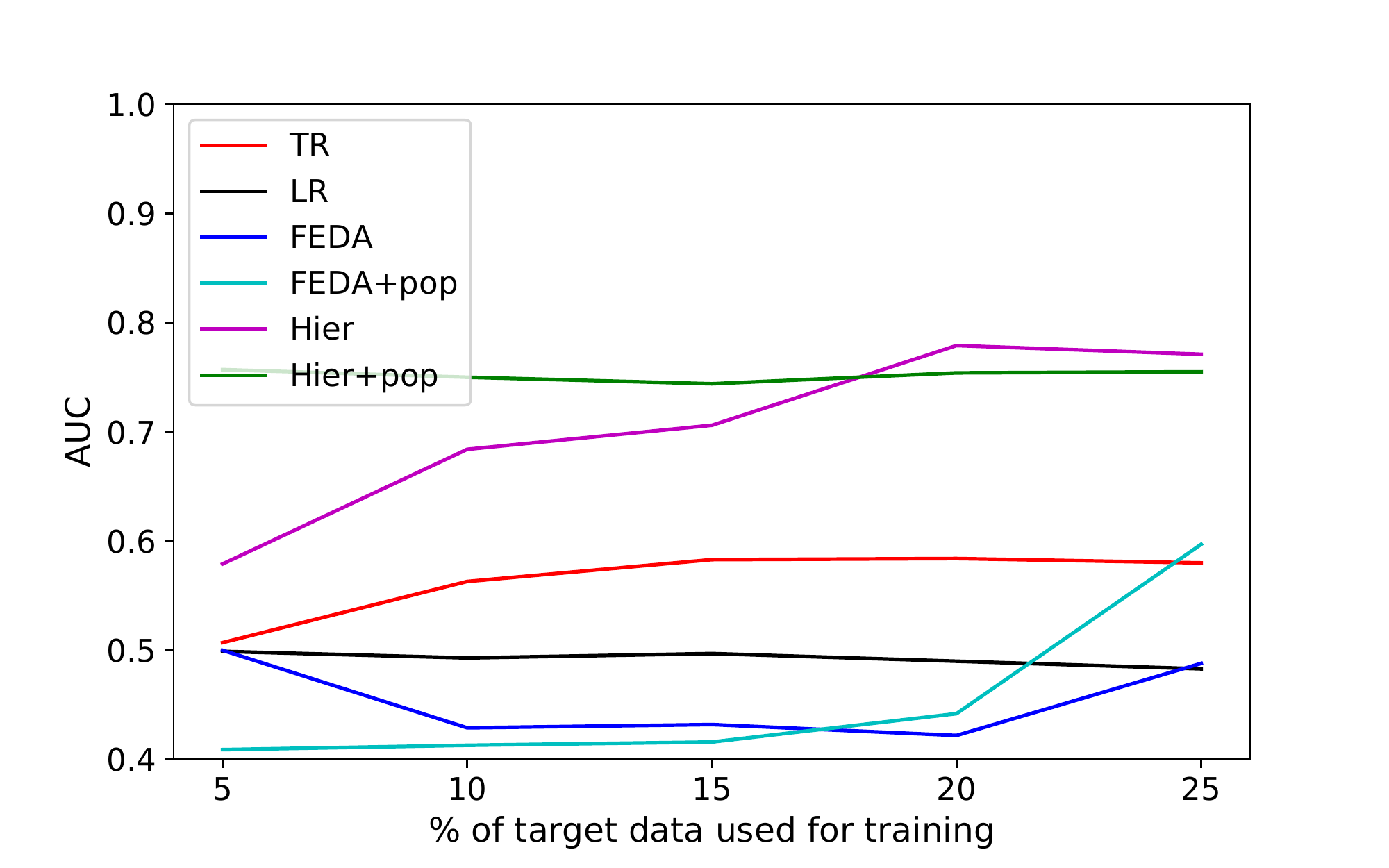}
    \caption{Fluwatch}
    \end{subfigure}\vspace{1pt}
    \begin{subfigure}{0.5\textwidth}
    \includegraphics[width=1.1\linewidth]{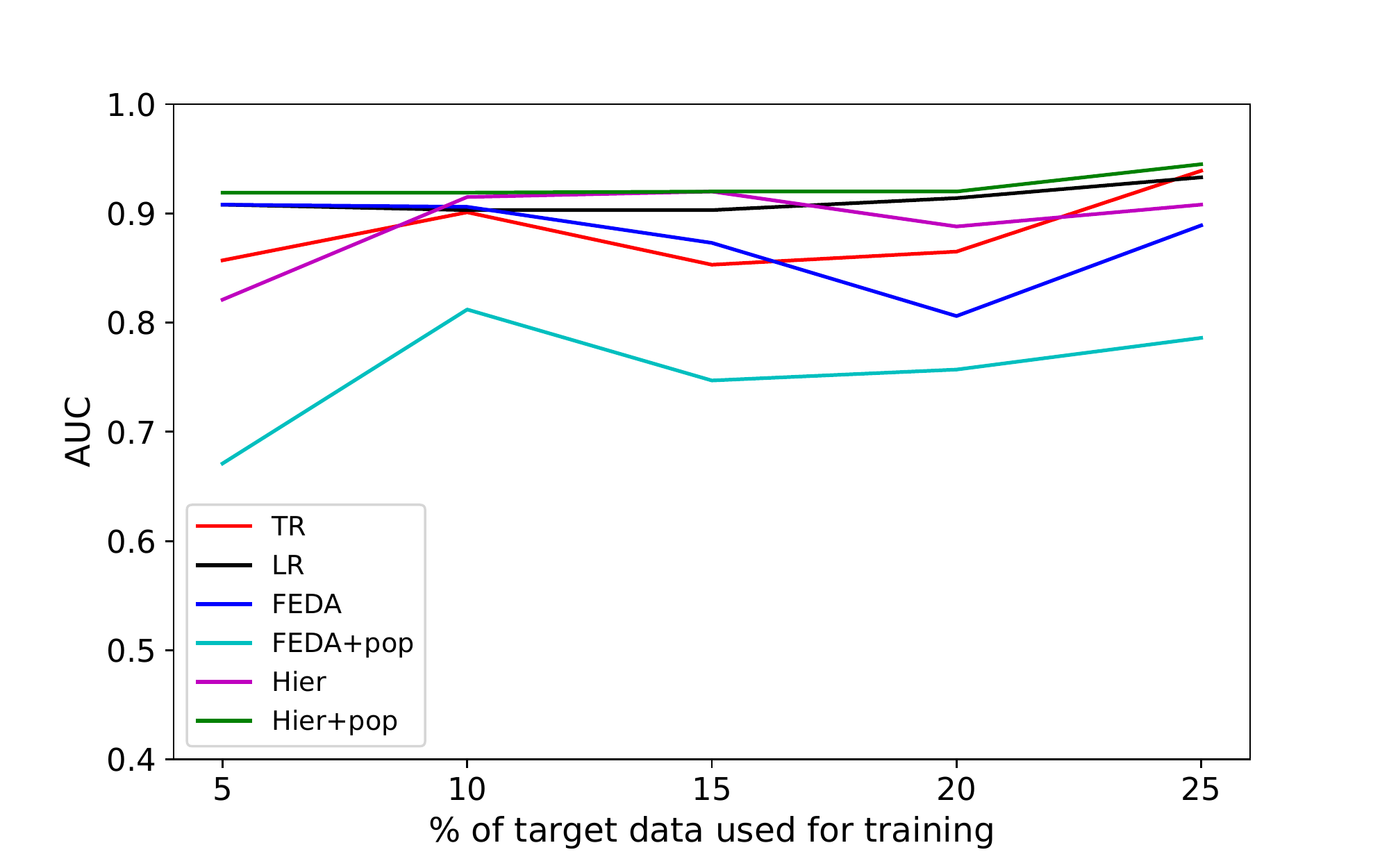}
    \caption{Hongkong}
    \end{subfigure}%
    \begin{subfigure}{0.5\textwidth}
    \includegraphics[width=1.1\linewidth]{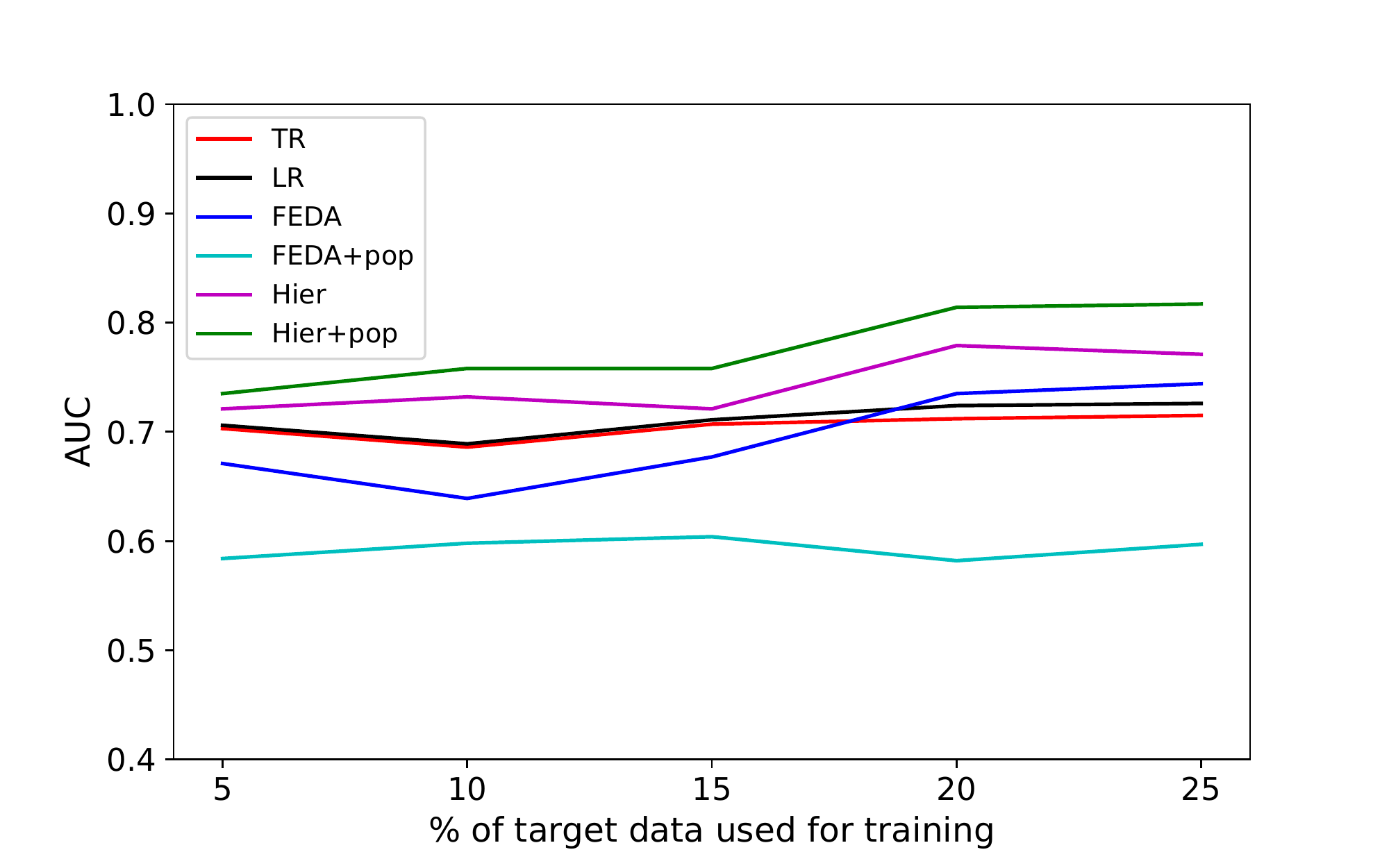}
    \caption{Hutterite}
    \end{subfigure}%
    
    \caption{Performance of Hier+pop method in comparison with baseline methods across increasing proportion of labelled target averaged across all population subgroups.}
    \label{fig:performance}
    
\end{figure*}

\begin{table*}[t]
\centering
 \caption {AUC scores across population subgroups ($\dagger$ denotes use of $\theta^l$ otherwise $\theta^d$ is used) with 20\% data used for training, bold values correspond to best performing model across population subgroup. Missing values (-) indicates the subgroup lacks data in either of the positive or negative classes, therefore models could not be trained for these subgroups.}
 \label{results:subgroup}

\begin{tabular}{l l|l l|l l|l l|l l |l l}
    \toprule
    \textbf{Dataset} & \textbf{Method} &
      \multicolumn{2}{c}{Age 0-5} &
      \multicolumn{2}{c}{Age 5-15} &
      \multicolumn{2}{c}{Age 16-44} & 
      \multicolumn{2}{c}{Age 45-64}&
      \multicolumn{2}{c}{Age 65+} \\

  &  & \text{Males} & \text{Females} & \text{Males} & \text{Females} & \text{Males} & \text{Females} & \text{Males} & \text{Females} &\text{Males} & \text{Females}  \\
    \midrule
& TR & \text{-} & \text{-} & \text{-} & \text{-} & 0.655 & 0.796 & 0.661 & 0.524 & 0.238 & 0.692 \\

 & LR & \text{-} & \text{-} & \text{-} & \text{-} & 0.541 & 0.845 & 0.774 & 0.607 & 0.188 & 0.542 \\

\textit{Goviral} & FEDA & \text{-} & \text{-} & \text{-} & \text{-} & 0.638 & 0.845 & 0.728 & 0.464 & 0.112 & \textbf{0.742} \\

& FEDA+pop & \text{-} & \text{-} & \text{-} & \text{-} & 0.524 & 0.602 & \textbf{0.807} & 0.214  & 0.112 & 0.700\\
 
& Hier & \text{-} & \text{-} & \text{-} & \text{-} & 0.681 & 0.767 & 0.766 & 0.645 & 0.312 & 0.700 \\ 

& Hier+pop & \text{-} & \text{-} & \text{-} & \text{-} & \textbf{0.842} & \textbf{0.910} & \textbf{0.807} & \textbf{0.666} & \textbf{0.500}{$\dagger$} &  \textbf{0.742}\\

\hline
    
& TR & \text{-} & 0.158 & 0.762 & 0.726 &  0.808 & 0.708 & 0.293 & 0.678 & 0.239 & 0.789 \\

 & LR & \text{-} & 0.440 & 0.286 & 0.583 & 0.295 & 0.467 & 0.384 & 0.486 & 0.647 & 0.588 \\

\textit{Fluwatch} & FEDA & \text{-} & 0.711 & 0.465 & 0.577 & 0.105 & 0.232 & 0.530 & 0.118 & 0.719 & 0.284 \\

& FEDA+pop & \text{-} & 0.440 & 0.298 & 0.565 & 0.311 & 0.412 & 0.389 & 0.260 & 0.803 & 0.505\\  

& Hier & \text{-} & 0.710 & 0.547 & 0.63 & 0.676 & 0.691 & 0.388 & \textbf{0.767} & 0.323 & \textbf{0.833} \\

& Hier+pop & \text{-} &  \textbf{0.868} & \textbf{0.747}{$\dagger$} &  \textbf{0.928}{$\dagger$} & \textbf{0.787}{$\dagger$} &  \textbf{0.757}{$\dagger$} &  \textbf{0.757} &  \textbf{0.767}{$\dagger$} &  \textbf{0.847} & \textbf{0.833}{$\dagger$}\\

\hline
& TR & \text{-} & 0.156 & 0.961 & 0.963 & 0.954 & 0.959 & 0.997 & 0.878 & 0.929 & \textbf{0.922}\\

& LR & \text{-} & 0.156 & 0.957 & 0.960 & 0.948 & 0.960 & 0.997 & 0.880 & 1.000 & \textbf{0.922}\\

\textit{Hongkong}  & FEDA & - & 0.156 & 0.936 & 0.943 & 0.921 & 0.873 & 0.957 & 0.810 & 0.864 & 0.797\\

& FEDA+pop & \text{-} & 0.156 & 0.957 & 0.710 & 0.948 & 0.752 & \textbf{0.997} & 0.674 & 0.773 & 0.583 \\

& Hier & \text{-} & 0.211 & 0.975 & 0.990 & 0.991 & 0.993 & 0.980 & 0.930 & \textbf{1.000} & \textbf{0.922} \\

& Hier+pop & \text{-} &  \textbf{0.500}{$\dagger$} &  \textbf{0.995} &  \textbf{1.000} &  \textbf{0.995} &  \textbf{0.996} &  \textbf{0.997} & \textbf{0.939} &  \textbf{1.000} &  \textbf{0.922} \\
\hline

& TR & 0.714 & 0.750 & 0.780 & 0.995 & 0.358 & 0.860 & 0.741 & 0.971 & 0.286 & 0.320\\

 & LR & 0.532 & 0.902 & 0.904 & 0.793 & 0.431 & 0.780 & 0.76 9  & 0.783 & 0.405 & 0.180\\

\textit{Hutterite} & FEDA & 0.532 & 0.902 & 0.904 & 0.793 & 0.431 & 0.780 & 0.769 & 0.783 & 0.405 & 0.180\\

& FEDA+pop & 0.500 & 0.500 & 0.671 & 0.604 & 0.380 & 0.847 & 0.834 & 0.906 & 0.405 & 0.180 \\

& Hier & 0.831 & \textbf{0.902} & \textbf{0.957} & 0.792 & 0.380 & 0.760 & 0.834 & 0.760 & 0.404 & 0.180 \\

& Hier+pop &  \textbf{0.851} &  \textbf{0.902}{$\dagger$} &  \textbf{0.957} & \textbf{1.000} &  \textbf{0.576}{$\dagger$} &  \textbf{1.000} &  \textbf{0.890} & \textbf{0.971} & \textbf{0.500} & \textbf{0.500}\\
\bottomrule

  \end{tabular}
  \label{tab:appendix_aucparam}
\end{table*}
As motivated, we consider the case of transferring information from multiple source data sets from different domains to a largely unlabelled target dataset. 
We conduct multiple experiments to compare the proposed framework with relevant baselines to specifically examine the value of i) the hierarchical structure and ii) incorporation of population attributes, and iii) the amount of labelled data available from the target. Area under the ROC curve (AUC) metric is used to assess the performance based on both sensitivity and specificity. AUC is a measure of goodness of the ability of a binary classifier, equal to the probability that the classifier will rank a randomly chosen positive instance higher than a randomly chosen negative one. We evaluate AUC across all the population subgroups of the dataset ($D_{a,g}$). We compare results to three methods: Target only (\textbf{TR}), Logistic Regression (\textbf{LR}), Frustratingly Easy Domain Adaptation, which is noted for extreme simplicity and was used previously on symptom data \cite{daume2009frustratingly,rehman2018domain}, without (\textbf{FEDA}) and with demographic attributes (\textbf{FEDA+pop}), Undirected Hierarchical Bayesian Domain adaptation without (\textbf{Hier}) and with demographic attributes (\textbf{Hier+pop})\footnote{Code is available at \url{https://github.com/ChunaraLab/Pop-aware-domain-adaptation}}. Based on how the methods are designed, we describe how training and testing must work for each. The \textit{Target only} method is trained only on a subset of the target environment dataset without incorporation of any information from source environments. \textit{Logistic Regression} is trained on all the source datasets and a subset of the target dataset. \textit{Frustratingly Easy Domain Adaptation} incorporates features specific to the source, specific to the target as well as a union of the source and target dataset, and is trained on both the source and subset of the target datasets. The proposed method, \textit{Hier+pop}, is trained on both the source datasets and a subset of the target. All methods are tested on the target dataset (excluding the subset of the target data used for training).

\section{Results, Discussion and Impact}

 This work is relevant to the increasing scenarios in which algorithms are being used to co-analyse and use multiple real-world datasets, gathered from different contexts and composed of different population distributions. We present a novel approach in the framework of observational transport, applicable in scenarios with instability in observed variables and selection bias; a significant challenge in many health transport problems. The model is motivated by knowledge of the underlying causal model. By testing on four real-world datasets for an influenza prediction from symptoms task, we show the multi-component model significantly improves performance by using principles of domain adaptation as well as by capturing information shared among population subgroups through a hierarchical and joint optimization approach. We perform a rigorous evaluation showing that with low amounts of labelled target data the model performs consistently better than baselines on entire datasets and on individual subgroups even when underrepresented in a specific dataset. 

\subsection{Performance analysis} Of the methods compared, TR and LR have the poorest performance (Table \ref{tab:result}) across entire datasets. This makes sense, as a target-only model doesn't incorporate any information from other environments or populations. And, LR doesn't account for any population attributes.  In all cases the Hier+pop method which accounts for the demographic attributes without including the demographic parameters explicitly in the same feature space as the symptoms (as is done by FEDA+pop), gives best performance across entire datasets. This also confirms the need to have different symptom parameters for specific demographic subgroups. We studied performance further based on amount of labelled training data available. We observe that Hier+pop performs consistently better than the baselines at low amounts of labelled target data (Figure \ref{fig:performance}). It should be noted that we examined results above 25\% labels, and trends continue. As more labelled data becomes available, TR improves substantially as expected. Comparing datasets, Goviral has limited sample size (Figure \ref{fig:demo}), leading to low performance of baseline methods, and Hier+pop captures the invariant information across the source environments to improve performance over baselines drastically. As compared to Goviral, Hutterite has better representation of the population subgroups and hence the baselines do not perform poorly but Hier+pop still performs substantially better. We highlight these results for Goviral and Hutterite datasets due to the vastly different sample sizes and data collection environment; results for other datasets follow the same trends (Figure \ref{fig:performance}). This demonstrates that multi-component invariant learning helps capture information shared among subgroups even when they are underrepresented. We also examined the learned parameters for the subgroups ($\mathcal{D}_{a,g}$), finding that they comply to conditions discussed in the Licensing conditions subsection. We also analyze performance of the methods by subgroup and find that Hier+pop indeed has better prediction across the subgroups, competing closely with TR in the case where $\theta^l$ are used instead of the invariant parameters $\theta^d$ (Table \ref{tab:appendix_aucparam}). In these specific cases, as expected, local information is preferred, therefore $\theta^l$ for the target dataset, which is influenced by the source environments, leads to a dip in the performance as compared to TR which does not have any influence by the source datasets. 

\subsection{Performance across population subgroups}
Performance across subgroups for Goviral, Fluwatch, Hongkong and Hutterite is reported in Table \ref{results:subgroup}. Hier+pop performs consistently better than the baselines across all the population subgroups for multiple datasets. We also report where the dataset specific parameters ($\theta^l$) are used instead of the invariant ($\theta^d$). This complies with the conditions provided in Theorem 1.\\

\subsection{Performance across increasing proportion of labelled target data}
Hier+pop performs consistently better than the baselines (TR, LR, FEDA, FEDA+pop) for multiple datasets. The trend is similar across the different datasets as shown in Figure \ref{fig:performance}. With increasing amounts of target data available, methods like Target only and FEDA perform well as more information about the dataset-specific feature space is available and there is less reliance on population-invariant information. In such scenarios the target dataset will have rich information about the target environment specific population subgroups.

\subsection{Relations to fairness and social sciences}
This work is motivated by real challenges health. We derive the proposed methodology in a principled manner from the causal structure of the problem. At the same time, there are overlaps with current research in fairness and the social sciences. First, formulation of the model in a hierarchical structure (versus with fixed effects) corroborates the latest thinking in social science theory, as it doesn't treat characteristics across multiple types of  parameters as additive \cite{evans2018multilevel}. Partitioning the variance across different aspects (symptoms and demographics), is meaningful as well, because the causes of variation at different levels may differ (e.g. causes of symptoms versus manifestations in different population subgroups). We derive this from the causal structure, but it also has been discussed in the epidemiology literature \cite{susser1996choosing}. Moreover, hierarchical structure allows understanding of different datasets simultaneously rather than making reference to one perfect or representative dataset (e.g. by one reference group for each variable and parameter).

As well, while the aim of this study isn't to ensure specific fairness criteria are fulfilled mainly due to multiple sensitive attributes, the work still addresses an important problem identified in the fairness community: population subgroups can be represented differently in different datasets. Since the subgroups are characterized by multiple sensitive attributes (age and gender), some non-binary (age), imposing fairness constraints like demographic parity is not suitable especially when characteristics are shared across demographic subgroups. However, research in the fairness and machine learning literature has discussed how a one-size fits all model for all population subgroups suffers from aggregation bias when the conditional distribution $P\left(Y \mid X\right)$ is not consistent across the subgroups as shown by \citet{suresh2019framework}. We show here, for an influenza prediction task, that models that do not incorporate any subgroup-level information are not optimal across all the subgroups and are fitted to the dominant population subgroup. Hier+pop on the other hand mitigates the issue of aggregation bias by incorporating information across multiple specific subgroups. 

One of the primary aims of formulating Hier+pop is thus to not compromise on sub-population AUC owing to selection bias and representation bias. We have drawn on principles such as beneficence (``do-the-best'') and non-maleficence (``do-no-harm'') and provide an approach to learn the parameters for each subgroup independently. As literature in fairness has discussed, model performance is often compromised due to inadequate sample sizes and appropriate data collection can aid in mitigating this issue \cite{chen2018my}.  \citet{ustun2019fairness} have developed an approach to learn decoupled classifiers for each subgroup of the population, thereby ensuring fairness across the subgroups without performing any harm. Our work, capturing information in a hierarchical manner also is towards ensuring that performance on lesser sized subgroups is not comprised. While in large healthcare systems representation in multiple subgroups can be reached, often individual healthcare institutions or epidemiological studies, as demonstrated here, only reach a certain population distribution, and sourcing new data is simply not feasible. Accordingly, our approach by combining subgroup information across multiple environments is pragmatic. While methods like FEDA+pop and Hier performed at par with Hier+pop for a few select subgroups with adequately representative samples, Hier+pop performs consistently well across all the subgroups, and all four datasets.

\subsection{Impact}
As new datasets are constantly being generated in different environments and from different constituent populations, the model and findings from this work can be used in multiple ways by those designing surveillance systems. For example, to proactively assess and inform which population subgroups need to be further sampled to improve prediction in the target data (by comparing $\theta^l$ and $\theta^d$ across datasets). Knowledge of the criteria regarding local versus invariant parameters can also be used to identify which datasets can be combined to improve prediction. Practically, this work demonstrates how practitioners can save effort and cost by only labeling a proportion of data and combining data with other datasets to improve prediction. Overall, we present a  practical approach to combine information from multiple instances especially when it is difficult to obtain labels; which can often be the case in public health, while retaining the global characteristics of population subgroups shared across environments.

\section{Acknowledgments}
This work was supported in part by grant 1845487 from the National Science Foundation.
\bibliographystyle{abbrvnat}
\bibliography{aaai20ref.bib}
\clearpage

\appendix


\noindent
Supplementary material
\section{Proof of Concept for Theorem 1}
\label{formal_proof}
We have two conditions:
\begin{enumerate}
    \item 
   $P\left(X=1 \mid Y=1,l\right) > P\left(X=0 \mid Y=1,l\right)$ \\ $\left( \text{let} \quad \delta_l = \delta_l^{+},\\P\left(X=1 \mid Y=1,l\right) = \delta_{l}^{+_{1}}, \\P\left(X=0 \mid Y=1,l\right) = \delta_{l}^{+_{0}}\right)$.
   \item 
   $P\left(X=1 \mid Y=1,l\right) < P\left(X=0 \mid Y=1,l\right)$ \\$\left( \text{let} \quad  \delta_l = \delta_l^{-},\\P\left(X=1 \mid Y=1,l\right) = \delta_{l}^{-_{1}},\\ P\left(X=0 \mid Y=1,l\right) = \delta_{l}^{-_{0}}\right)$.
    
\end{enumerate}

\noindent Let us consider $\delta_{\mathcal{D}} > \delta_{\mathit{pop}}$; we encounter four conditions :
\begin{enumerate}
    \item $\delta_{\mathcal{D}} = \delta_{\mathcal{D}}^{+}$, $\delta_{\mathit{pop}} = \delta_{\mathit{pop}}^{+}$ then 
    $\delta_{\mathcal{D}}^{+_{1}} - \delta_{\mathcal{D}}^{+_{0}} >\delta_{\mathit{pop}}^{+_{1}} - \delta_{\mathit{pop}}^{+_{0}} $\\
    $\delta_{\mathcal{D}}^{+_{0}} = 1-\delta_{\mathcal{D}}^{+_{1}}$, similarly $\delta_{\mathit{pop}}^{+_{0}} = 1-\delta_{\mathit{pop}}^{+_{1}}$. \\ Therefore, $2\delta_{\mathcal{D}}^{+_{1}} -1 >2\delta_{\mathit{pop}}^{+_{1}} -1 $\\ $\implies P\left(X=1 \mid Y=1,l=\mathcal{D}\right) > P\left(X=1 \mid Y=1,l=pop\right) $ and $I_\mathcal{D} < I_{\mathit{pop}}$.
    
    \item $\delta_{\mathcal{D}} = \delta_{\mathcal{D}}^{+}$, $\delta_{\mathit{pop}} = \delta_{\mathit{pop}}^{-}$ then \\$(\delta_{\mathcal{D}}^{+_{1}} - \delta_{\mathcal{D}}^{+_{0}}) + (\delta_{\mathit{pop}}^{+_{1}} - \delta_{\mathit{pop}}^{+_{0}}) > 0$ and $(\delta_{\mathcal{D}}^{+_{1}} - \delta_{\mathcal{D}}^{+_{0}}) > 0, (\delta_{\mathit{pop}}^{+_{1}} - \delta_{\mathit{pop}}^{+_{0}}) < 0) \implies \delta_{\mathcal{D}}^{+_{1}} - \delta_{\mathcal{D}}^{+_{0}} >\delta_{\mathit{pop}}^{+_{1}} - \delta_{\mathit{pop}}^{+_{0}} $; \\
    which results into \\
    $P\left (X=1 \mid Y=1,l=\mathcal{D}\right) > P\left (X=1 \mid Y=1,l=\mathit{pop} \right) $ \\
    and $I_\mathcal{D} < I_{\mathit{pop}}$.
    
    \item $\delta_{\mathcal{D}} = \delta_{\mathcal{D}}^{-}$, $\delta_{\mathit{pop}} = \delta_{\mathit{pop}}^{+}$ then \\$(\delta_{\mathcal{D}}^{+_{1}} - \delta_{\mathcal{D}}^{+_{0}}) + (\delta_{\mathit{pop}}^{+_{1}} - \delta_{\mathit{pop}}^{+_{0}}) < 0$ and $(\delta_{\mathcal{D}}^{+_{1}} - \delta_{\mathcal{D}}^{+_{0}}) < 0, (\delta_{\mathit{pop}}^{+_{1}} - \delta_{\mathit{pop}}^{+_{0}}) > 0) \implies \delta_{\mathcal{D}}^{+_{1}} - \delta_{\mathcal{D}}^{+_{0}} >\delta_{\mathit{pop}}^{+_{1}} - \delta_{\mathit{pop}}^{+_{0}} $; \\
    which results into \\
    $P\left(X=1 \mid Y=1,l=\mathcal{D}\right) > P\left(X=1 \mid Y=1,l=\mathit{pop}\right) $ \\ 
    and $I_\mathcal{D} < I_{\mathit{pop}}$.
    
    \item $\delta_{\mathcal{D}} = \delta_{\mathcal{D}}^{-}$, $\delta_{\mathit{pop}} = \delta_{\mathit{pop}}^{-}$ then \\$(\delta_{\mathcal{D}}^{-_{1}} - \delta_{\mathcal{D}}^{-_{0}}) - (\delta_{\mathit{pop}}^{-_{1}} - \delta_{\mathit{pop}}^{-_{0}}) > 0 \implies \delta_{\mathcal{D}}^{-_{1}} - \delta_{\mathcal{D}}^{-_{0}} >\delta_{\mathit{pop}}^{-_{1}} - \delta_{\mathit{pop}}^{-_{0}} $;\\ 
    which results into \\
    $P\left(X=1 \mid Y=1,l=\mathcal{D}\right) > P\left(X=1 \mid Y=1,l=\mathit{pop}\right) $ \\
    and $I_\mathcal{D} < I_{\mathit{pop}}$.
 \end{enumerate}

\noindent For all of the conditions we obtain that $I_\mathcal{D} < I_{pop}$ which means that there is more information available from the entire population from all the environments rather than just the specific dataset.


\end{document}